\documentclass[10pt,twocolumn,letterpaper]{article}

\usepackage{cvpr}              

\usepackage{graphicx}
\usepackage{amsmath}
\usepackage{amssymb}
\usepackage{booktabs}

\usepackage{algorithm}
\usepackage{algorithmic}
\usepackage{booktabs}
\usepackage{tabularx}
\usepackage{multirow}

\usepackage[pagebackref,breaklinks,colorlinks]{hyperref}

\usepackage[capitalize]{cleveref}
\crefname{section}{Sec.}{Secs.}
\Crefname{section}{Section}{Sections}
\Crefname{table}{Table}{Tables}
\crefname{table}{Tab.}{Tabs.}

\def\cvprPaperID{7876} 
\def\confName{CVPR}
\def\confYear{2022}

\begin{document}

\title{Dual Cross-Attention Learning for Fine-Grained Visual Categorization and Object Re-Identification}


\author{Haowei Zhu\footnotemark[1], Wenjing Ke\footnotemark[1], Dong Li, Ji Liu, Lu Tian, Yi Shan \\
 Advanced Micro Devices, Inc., Beijing, China \\
{\tt\small \{haowei.zhu, wenjing.ke, d.li, lu.tian, yi.shan\}@amd.com}
}

\maketitle

\renewcommand{\thefootnote}{\fnsymbol{footnote}}
\footnotetext[1]{Equal contribution.}

\begin{abstract}
Recently, self-attention mechanisms have shown impressive performance in various NLP and CV tasks, which can help capture sequential characteristics and derive global information. In this work, we explore how to extend self-attention modules to better learn subtle feature embeddings for recognizing fine-grained objects, e.g., different bird species or person identities. To this end, we propose a dual cross-attention learning (DCAL) algorithm to coordinate with self-attention learning. First, we propose global-local cross-attention (GLCA) to enhance the interactions between global images and local high-response regions, which can help reinforce the spatial-wise discriminative clues for recognition. Second, we propose pair-wise cross-attention (PWCA) to establish the interactions between image pairs. PWCA can regularize the attention learning of an image by treating another image as distractor and will be removed during inference. We observe that DCAL can reduce misleading attentions and diffuse the attention response to discover more complementary parts for recognition. We conduct extensive evaluations on fine-grained visual categorization and object re-identification. Experiments demonstrate that DCAL performs on par with state-of-the-art methods and consistently improves multiple self-attention baselines, e.g., surpassing DeiT-Tiny and ViT-Base by 2.8\% and 2.4\% mAP on MSMT17, respectively.
\end{abstract}


\section{Introduction}

Self-attention is an attention mechanism that can relate different positions of a single sequence and draw global dependencies. It is originally applied in natural language processing (NLP) tasks \cite{vaswani2017attention,devlin2018bert} and exhibits the outstanding performance. Recently, Transformer with self-attention learning has also been explored for various vision tasks (e.g., image classification \cite{dosovitskiy2020image,chen2020generative,touvron2020training,ramachandran2019stand,hu2019lrnet,wang2020axial} and object detection \cite{carion2020detr,zhu2020deformable}) as an alternative of convolutional neural network (CNN). For general image classification, self-attention has been proved to work well for recognizing 2D images by viewing image patches as words and flattening them as sequences \cite{dosovitskiy2020image,touvron2020training}.

In this work, we investigate how to extend self-attention modules to better learn subtle feature embeddings for recognizing fine-grained objects, e.g., different bird species or person identities. Fine-grained recognition is more challenging than general image classification owing to the subtle visual variations among different sub-classes. Most of existing approaches build upon CNN to predict class probabilities or measure feature distances. To address the subtle appearance variations, local characteristics are often captured by learning spatial attention \cite{fu2017look,zheng2017learning,sun2018multi,luo2019cross} or explicitly localizing semantic objects / parts \cite{zheng2019looking,ding2019selective,yang2018learning, zhang2019learning}.
We adopt a different way to incorporate local information based on vision Transformer. To this end, we propose global-local cross-attention (GLCA) to enhance the interactions between global images and local high-response regions. Specifically, we compute the cross-attention between a selected subset of query vectors and the entire set of key-value vectors. By coordinating with self-attention learning, GLCA can help reinforce the spatial-wise discriminative clues to recognize fine-grained objects.

Apart from incorporating local information, another solution to distinguish the sutble visual differences is pair-wise learning. The intuition is that one can identify the subtle variations by comparing image pairs. Exiting CNN-based methods design dedicated network architectures to enable pair-wise feature interaction \cite{zhuang2020learning,gao2020channel}. A contrastive loss \cite{gao2020channel} or score ranking loss \cite{zhuang2020learning} is used for feature learning. Motivated by this, we also employ a pair-wise learning scheme to establish the interactions between image pairs. Different from optimizing the feature distance, we propose pair-wise cross-attention (PWCA) to regularize the attention learning of an image by treating another image as distractor. Specifically, we compute the cross-attention between query of an image and combined key-value from both images. By introducing confusion in key and value vectors, the attention scores are diffused to another image so that the difficulty of the attention learning of the current image increases. Such regularization allows the network to discover more discriminative regions and alleviate overfitting to sample-specific features. It is noted that PWCA is only used for training and thus does not introduce extra computation cost during inference. 

The proposed two types of cross-attention are easy-to-implement and compatible with self-attention learning. We conduct extensive evaluations on both fine-grained visual categorization (FGVC) and object re-identification (Re-ID). Experiments demonstrate that DCAL performs on par with state-of-the-art methods and consistently improves multiple self-attention baselines. Particularly, for FGVC, DCAL improves DeiT-Tiny by 2.5\% and reaches 92.0\% top-1 accuracy with the larger R50-ViT-Base backbone on CUB-200-2011. For Re-ID, DCAL improves DeiT-Tiny and ViT-Base by 2.8\% and 2.4\% mAP on MSMT17, respectively.

Our main contributions can be summarized as follows. (1) We propose global-local cross-attention to enhance the interactions between global images and local high-response regions for reinforcing the spatial-wise discriminative clues. (2) We propose pair-wise cross-attention to establish the interactions between image pairs by regularizing the attention learning. (3) The proposed dual cross-attention learning can complement the self-attention learning and achieves consistent performance improvements over multiple vision Transformer baselines on various FGVC and Re-ID benchmarks.

\section{Related Work}

\subsection{Self-Attention Mechanism}

The self-attention mechanism is originally proposed to relate distinct positions in a sequence and draw global dependencies. Transformer carrying forward this mechanism has dominated in various sequence-to-sequence NLP tasks \cite{vaswani2017attention,devlin2018bert}. Transformer usually consists of multiple encoder and decoder modules. Each encoder / decoder includes a multi-head self-attention (MSA) layer and a feed-forward network (FFN) layer. A decoder also has an extra MSA layer to handle the output of encoder. Besides, layer normalization (LN) and residual connection are used in each MSA or FFN layer. Recent work has applied Transformers to various vision tasks (e.g., image classification \cite{dosovitskiy2020image,chen2020generative,touvron2020training,ramachandran2019stand,hu2019lrnet,wang2020axial}, object detection \cite{carion2020detr,sun2020tsp,sun2020sparse,zhu2020deformable}, semantic segmentation \cite{zheng2020rethinking,huang2019ccnet,wang2020axial,srinivas2021botnet,wang2020max} and low-level tasks \cite{chen2020pre}) and shown competitive performance compared to the state-of-the-art CNNs. For general image classification, iGPT \cite{chen2020generative} first uses auto-regressive and BERT \cite{devlin2018bert} objectives for self-supervised pre-training and then fine-tunes for classification tasks. ViT \cite{dosovitskiy2020image} reshapes an image into a sequence of flattened fixed-size patches for training Transformer encoders only. Attempts have also been made to improve ViT by knowledge distillation \cite{touvron2020training} and progressive tokenization \cite{yuan2021tokens}.
Fine-grained recognition is more challenging than general image classification owing to the sutble visual variations among different sub-classes. In this work, we extend self-attention to better recognize fine-grained objects with two types of cross-attention modules.

\subsection{Fine-Grained Visual Categorization}
Fine-grained visual categorization (FGVC) is a special case of image classification, which aims to identify those highly-confused categories with fine differences. Prior CNN-based methods address this task by mining effective information from multi-level features \cite{du2020fine,luo2019cross,zhang2019learning}, adopting multi-granularity training strategies \cite{du2020fine}, locating discriminative objects or parts \cite{zheng2019looking,ding2019selective} and exploring feature interaction in pair-wise learning \cite{zhuang2020learning,gao2020channel}. Recently, a few Transformer-based methods address FGVC by feature fusion on multi-level Transformer layers \cite{wang2021feature} and part selection \cite{he2021transfg}. Our motivation is similar with \cite{wang2021feature,he2021transfg} in the aspects of aggregating multi-level attention and selecting patch tokens. However, they are based on self-attention only while we design two cross-attention modules for learning. 

\subsection{Object Re-Identification} Similar to FGVC, object re-identification also aims to distinguish different person / vehicle identities with subtle inter-class differences. Mainstream Re-ID methods are based on the CNN structure and metric learning \cite {luo2019bag,liu2017end}. Local information is crucial for Re-ID and many different approaches have been presented by encoding discriminative part-level features \cite{sun2018beyond,wang2018learning,liu2020beyond}. Transformer with self-attention structure has recently been applied to Re-ID by introducing part tokens \cite{zhu2021aaformer}, shuffling patch embeddings \cite{he2021transfg}, and learning disentangled features \cite{jia2021drl}. Our work differs from the most related methods \cite{zhu2021aaformer,he2021transfg} in the following aspects. First, we adopt a different way to encode the local information by GLCA, while \cite{he2021transfg} does not explicitly mine part regions and \cite{zhu2021aaformer} computes the attention between a part token and its associated subset of patch embeddings by online clustering. Second, \cite{he2021transfg,zhu2021aaformer} uses a single image for training while we employ image pairs for PWCA. Third, \cite{he2021transfg} requires side information (e.g., camera IDs and viewpoint labels) while our method only takes images as input.

\section{Proposed Approach}
\label{sec:method}

\begin{figure*}[t!]
\footnotesize
\begin{center}
\begin{tabular}{@{}cc@{}}
\includegraphics[width = 0.4\linewidth]{{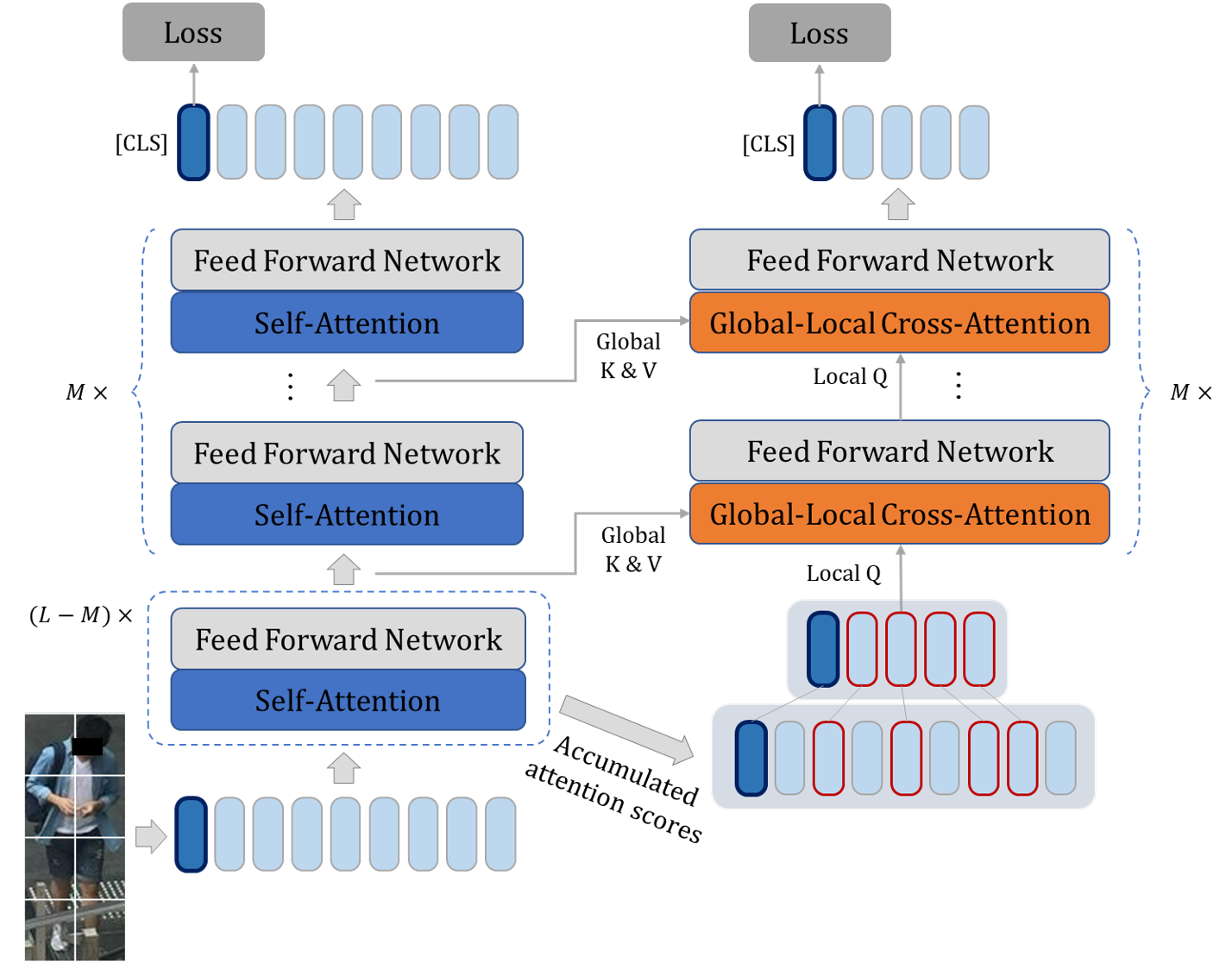}} & 
\includegraphics[width = 0.4\linewidth]{{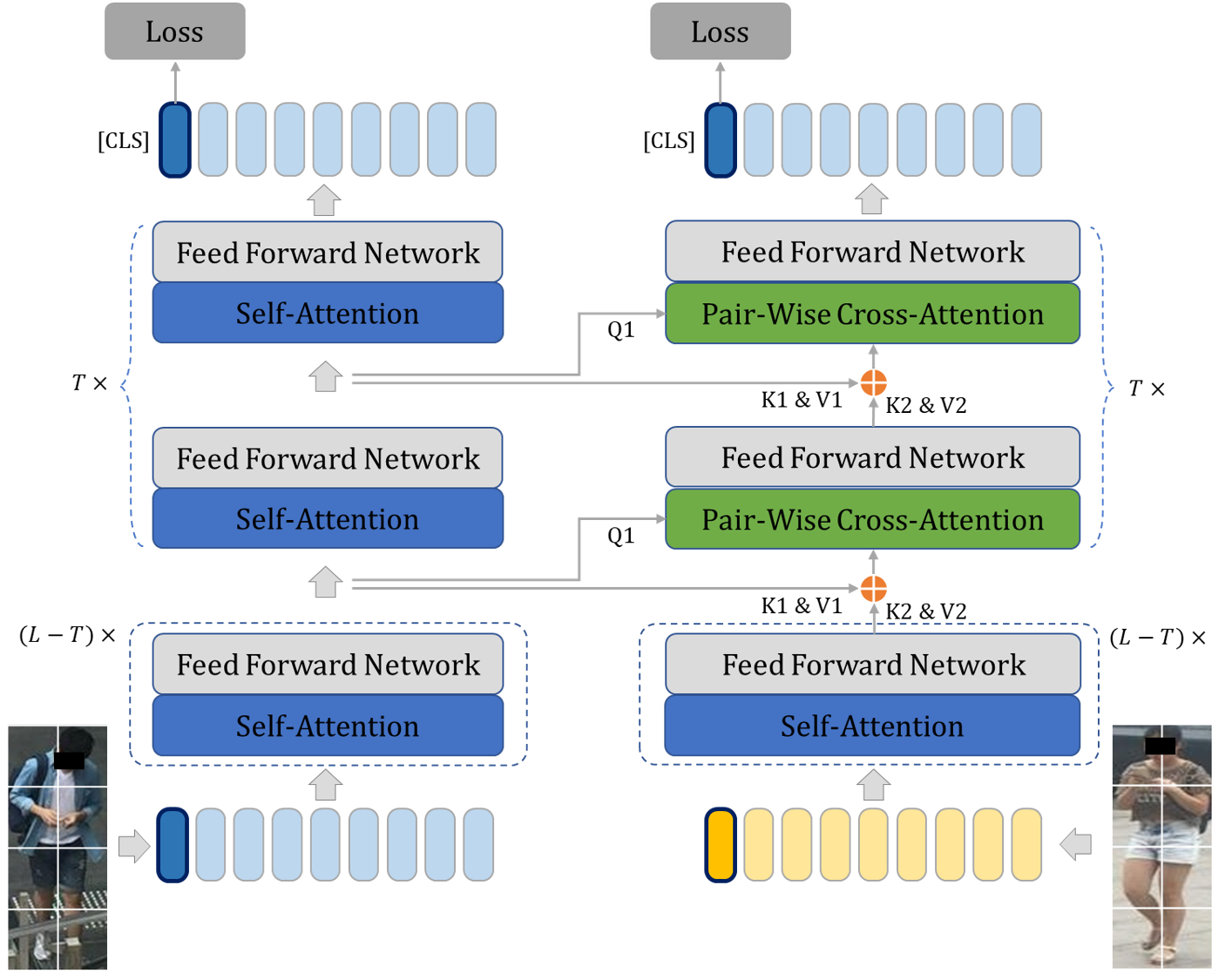}} \\
{\footnotesize (a) Global-Local Cross-Attention (GLCA)} & 
{\footnotesize (b) Pair-Wise Cross-Attention (PWCA)}
\end{tabular}
\end{center}
\vspace{-5mm}
\caption{Overview of the proposed two types of cross-attention mechanisms. We stack $L$ self-attention, $M$ global-local cross-attention, $T$ pair-wise cross-attention modules in our network. See Section \ref{sec:method} for details.}
\label{figure:overview}
\end{figure*}

\subsection{Revisit Self-Attention}

\cite{vaswani2017attention} originally proposes the self-attention mechanism to address NLP tasks by calculating the correlation between each word and all the other words in the sentence. \cite{dosovitskiy2020image} inherits the idea by taking each patch in the image / feature map as a word for general image classification. In general, a self-attention function can be depicted as mapping a query vector and a set of key and value vectors to an output. The output is computed as a weighted sum of value vectors, where the weight assigned to each value is computed by a scaled inner product of the query with the corresponding key. Specifically, a query $q \in \mathbb{R}^{1\times d}$ is first matched against $N$ key vectors ($K=[k_1;k_2;\cdots ;k_N]$, where each $k_i \in \mathbb{R}^{1\times d}$) using inner product. The products are then scaled and normalized by a softmax function to obtain $N$ attention weights. The final output is the weighted sum of $N$ value vectors ($V=[v_1;v_2;\cdots ;v_N]$, where each $v_i \in \mathbb{R}^{1\times d}$). By packing $N$ query vector into a matrix $Q=[q_1;q_2;\cdots ;q_N]$, the output matrix of self-attention (SA) can be represented as:
\begin{equation}
    f_{\mbox{SA}}(Q,K,V)\ =\mbox{softmax}(\frac{QK^T}{\sqrt{d}})V = SV
\label{eq:sa}
\end{equation}
where $\frac{1}{\sqrt{d}}$ is a scaling factor.  Query, key and value matrices are computed from the same input embedding $X \in \mathbb{R}^{N\times D}$ with different linear transformations: $Q=XW_Q$, $K=XW_K$, $V = XW_V$, respectively. $S \in \mathbb{R}^{N\times N}$ denotes the attention weight matrix. 

To jointly attend to information from different representation subspaces at different positions, multi-head self-attention (MSA) is defined by considering multiple attention heads. The process of MSA can be computed as linear transformation on the concatenations of self-attention blocks with subembeddings. To encode positional information, fixed / learnable position embeddings are added to patch embeddings and then fed to the network. To predict the class, an extra class embedding $\hat{\texttt{CLS}} \in \mathbb{R}^{1\times d}$ is prepended to the input embedding $X$ throughout the network, and finally projected with a linear classifer layer for prediction. Thus, the input embeddings as well as query, key and value matrices become $(N+1)\times d$ and the self-attention function (Eq. \ref{eq:sa}) allows to spread information between patch and class embeddings. 

Based on self-attention, a Transformer encoder block can be constructed by an MSA layer and a feed forward network (FFN). FFN consists of two linear transformation with a GELU activation. Layer normalization (LN) is put prior to each MSA and FFN layer and residual connections are used for both layers.

\subsection{Global-Local Cross-Attention} 
Self-attention treats each query equally to compute global attention scores according to Eq. \ref{eq:sa}. In other words, each local position of image is interacted with all the positions in the same manner. For recognizing fine-grained objects, we expect to mine discriminative local information to facilitate the learning of subtle features. To this end, we propose global-local cross-attention to emphasize the interaction between global images and local high-response regions. First, we follow attention rollout \cite{abnar2020quantifying} to calculate the accumulated attention scores for $i$-th block:
\begin{equation}
    \hat{S}_i = \bar{S}_i \otimes \bar{S}_{i-1} \cdots \otimes \bar{S}_1
    \label{eq:rollout}
\end{equation}
where $\bar{S}=0.5S+0.5E$ means the re-normalized attention weights using an identity matrix $E$ to consider residual
connections, $\otimes$ means the matrix multiplication operation. In this way, we track down the information propagated
from the input layer to a higher layer. Then, we use the aggregated attention map to mine the high-response regions. According to Eq. \ref{eq:rollout}, the first row of $\hat{S}_i = [\hat{s}_{i,j}]_{(N+1)\times (N+1)}$ means the accumulated weights of class embedding $\hat{\texttt{CLS}}$. We select top $R$ query vectors from $Q_i$ that correspond to the top $R$ highest responses in the accumulated weights of $\hat{\texttt{CLS}}$ to construct a new query matrix $Q^l$, representing the most attentive local embeddings. Finally, we compute the cross attention between the selected local query and the global set of key-value pairs as below.
\begin{equation}
f_{\mbox{GLCA}}(Q^l,K^g,V^g)=\mbox{softmax}(\frac{Q^l{K^g}^T}{\sqrt{d}})V^g 
\label{eq:glca}
\end{equation}

In self-attention (Eq. \ref{eq:sa}), all the query vectors will be interacted with the key-value vectors. In our GLCA (Eq. \ref{eq:glca}), only a subset of query vectors will be interacted with the key-value vectors. We observe that GLCA can help reinforce the spatial-wise discriminative clues to promote recognition of fine-grained classes. Another possible choice is to compute the self-attention between local query $Q^l$ and local key-value vectors ($K^l$, $V^l$). However, through establishing the interaction between local query and global key-value vectors, we can relate the high-response regions with not only themselves but also with other context outside of them. Figure \ref{figure:overview} (a) illustrates the proposed global-local cross-attention and we use $M=1$ GLCA block in our method.


\subsection{Pair-Wise Cross-Attention}
The scale of fine-grained recognition datasets is usually not as large as that of general image classification, e.g., ImageNet \cite{deng2009imagenet} contains over 1 million images of 1,000 classes while CUB \cite{wah2011caltech} contains only 5,994 images of 200 classes for training. Moreover, smaller visual differences between classes exist in FGVC and Re-ID compared to large-scale classification tasks. Fewer samples per class may lead to network overfitting to sample-specific features for distinguishing visually confusing classes in order to minimize the training error. 

To alleviate the problem, we propose pair-wise cross attention to establish the interactions between image pairs. PWCA can be viewed as a novel regularization method to regularize the attention learning. Specifically, we randomly sample two images ($I_1$, $I_2$) from the same training set to construct the pair. The query, key and value vectors are separately computed for both images of a pair. For training $I_1$, we concatenate the key and value matrices of both images, and then compute the attention between the query of the target image and the combined key-value pairs as follows:
\begin{equation}
    f_{\mbox{PWCA}}(Q_1,K_c,V_c) =\mbox{softmax}(\frac{Q_1 K_c^T}{\sqrt{d}})V_c
\label{eq:pwca}
\end{equation}
where $K_c=[K_1;K_2] \in \mathbb{R}^{(2N+2)\times d}$ and $V_c=[V_1;V_2] \in \mathbb{R}^{(2N+2)\times d}$. For a specific query from $I_1$, we compute $N+1$ self-attention scores within itself and $N+1$ cross-attention scores with $I_2$ according to Eq. \ref{eq:pwca}. All the $2N+2$ attention scores are normalized by the softmax function together and thereby contaminated attention scores for the target image $I_1$ are learned. 
Optimizing this noisy attention output increases the difficulty of network training and reduces the overfitting to sample-specific features. Figure \ref{figure:overview} (b) illustrates the proposed pair-wise cross-attention and we use $T=12$ PWCA blocks in our method. Note that PWCA is only used for training and will be removed for inference without consuming extra computation cost.


\section{Experiments}

\subsection{Experimental Setting}
\textbf{Datasets.}
We conduct extensive experiments on two fine-grained recognition tasks: fine-grained visual categorization (FGVC) and object re-identification (Re-ID). For FGVC, we use three standard benchmarks for evaluations: CUB-200-2011 \cite{wah2011caltech}, Stanford Cars \cite{krause20133d}, FGVC-Aircraft \cite{maji2013fine}.
For Re-ID, we use four standard benchmarks: Market1501 \cite{zheng2015scalable}, DukeMTMC-ReID \cite{wu2020deep}, MSMT17 \cite{wei2018person} for Person Re-ID and VeRi-776 \cite{zheng2020vehiclenet} for Vehicle Re-ID. In all experiments, we use the official train and validation splits for evaluation.

\textbf{Baselines.} 
We use DeiT and ViT as our self-attention baselines. In detail, ViT backbones are pre-trained on ImageNet-21k \cite{deng2009imagenet} and DeiT backbones are pre-trained on ImageNet-1k \cite{deng2009imagenet}. We use multiple architectures of DeiT-T/16, DeiT-S/16, DeiT-B/16, ViT-B/16, R50-ViT-B/16 with $L=12$ SA blocks for evaluation.

\textbf{Implementation Details.} 
We coordinate the proposed two types of cross-attention with self-attention in the form of multi-task learning. We build $L=12$ SA blocks, $M=1$ GLCA blocks and $T=12$ PWCA blocks as the overall architecture for training. The PWCA branch shares weights with the SA branch while GLCA does not share weights with SA. We follow \cite{zhang2021fairmot} to adopt dynamic loss weights for collaborative optimization, avoiding exhausting manual hyper-parameter search. The PWCA branch has the same GT target as the SA branch since we treat another image as distractor.

For FGVC, we resize the original image into 550$\times$550 and randomly crop to 448$\times$448 for training. The sequence length of input embeddings for self-attention baseline is $28\times 28=784$. We select input embeddings with top $R=10\%$ highest attention responses as local queries. We apply stochastic depth \cite{huang2016deep} and use Adam optimizer with weight decay of 0.05 for training. The learning rate is initialized as ${\rm lr}_{scaled}=\frac{5e-4}{512}\times batchsize$ and decayed with a cosine policy. We train the network for 100 epochs with batch size of 16 using the standard cross-entropy loss. 

For Re-ID, we resize the image into 256$\times$128 for pedestrian datasets, and 256$\times$256 for vehicle datasets. We select input embeddings with top $R=30\%$ highest attention responses as local queries. We use SGD optimizer with a momentum of 0.9 and a weight decay of 1e-4. The batch size is set to 64 with 4 images per ID. The learning rate is initialized as 0.008 and decayed with a cosine policy. We train the network for 120 epochs using the cross-entropy and triplet losses.

All of our experiments are conducted on PyTorch with Nvidia Tesla V100 GPUs. Our method costs 3.8 hours with DeiT-Tiny backbone for training using 4 GPUs on CUB, and 9.5 hours with ViT-Base for training using 1 GPU on MSMT17. During inference, we remove all the PWCA modules and only use the SA and GLCA modules. We add class probabilities output by classifiers of SA and GLCA for prediction for FGVC, and concat two final class tokens of SA and GLCA for prediction for Re-ID. A single image with the same input size as training is used for test.


\begin{table}[t!]
\begin{center}
\scalebox{0.9}{
\begin{tabular}{l|c|ccc}

\toprule
\multirow{2}{*}{Method} & \multirow{2}{*}{Backbone}        & \multicolumn{3}{c}{Accuracy (\%)}  \\
\multicolumn{1}{c|}{}   &   & \multicolumn{1}{c}{CUB} & \multicolumn{1}{c}{CAR} & \multicolumn{1}{c}{AIR}  \\ \midrule
RA-CNN \cite{fu2017look}                                       & VGG19                            & \multicolumn{1}{c}{85.3}              & \multicolumn{1}{c}{92.5}              & \multicolumn{1}{c}{88.4}                          \\
MA-CNN \cite{zheng2017learning}                                       & VGG19                            & \multicolumn{1}{c}{86.5}              & \multicolumn{1}{c}{92.8}              & \multicolumn{1}{c}{89.9}                             \\
MAMC \cite{sun2018multi}                                         & ResNet101                        & \multicolumn{1}{c}{86.5}              & \multicolumn{1}{c}{93.0}              & \multicolumn{1}{c}{-}                              \\
PC \cite{dubey2018pairwise}                                           & DenseNet161               & \multicolumn{1}{c}{86.9}             & \multicolumn{1}{c}{92.9}             & \multicolumn{1}{c}{89.2}                          \\
FDL \cite{liu2020filtration}                                          & DenseNet161                      & \multicolumn{1}{c}{89.1}             & \multicolumn{1}{c}{94.0}             & \multicolumn{1}{c}{-}                            \\
NTS-Net \cite{yang2018learning}                                      & ResNet50                         & \multicolumn{1}{c}{87.5}              & \multicolumn{1}{c}{93.9}              & \multicolumn{1}{c}{91.4}                          \\
Cross-X \cite{luo2019cross}                                     & ResNet50                         & \multicolumn{1}{c}{87.7}              & \multicolumn{1}{c}{94.6}              & \multicolumn{1}{c}{-}                             \\
S3N \cite{ding2019selective}                                          & ResNet50                         & \multicolumn{1}{c}{88.5}              & \multicolumn{1}{c}{94.7}              & \multicolumn{1}{c}{92.8}                          \\
MGE-CNN \cite{zhang2019learning}                                      & ResNet50                         & \multicolumn{1}{c}{88.5}              & \multicolumn{1}{c}{93.9}              & \multicolumn{1}{c}{-}                               \\
DCL \cite{chen2019destruction}                                          & ResNet50                         & \multicolumn{1}{c}{87.8}              & \multicolumn{1}{c}{94.5}              & \multicolumn{1}{c}{93.0}                             \\
TASN \cite{zheng2019looking}                                         & Resnet50                         & \multicolumn{1}{c}{87.9}              & \multicolumn{1}{c}{93.8}              & \multicolumn{1}{c}{-}                           \\
PMG \cite{du2020fine}                                          & ResNet50                         & \multicolumn{1}{c}{89.6}              & \multicolumn{1}{c}{95.1}              & \multicolumn{1}{c}{93.4}                          \\
CIN \cite{gao2020channel}   & ResNet50 & \multicolumn{1}{c}{88.1}  & \multicolumn{1}{c}{94.5}   & \multicolumn{1}{c}{92.8}   \\
API-Net\cite{zhuang2020learning}                                      & DenseNet161                      & \multicolumn{1}{c}{90.0}              & \multicolumn{1}{c}{95.3}              & \multicolumn{1}{c}{93.9}                          \\
LIO \cite{zhou2020look}                                          & ResNet50                         & \multicolumn{1}{c}{88.0}              & \multicolumn{1}{c}{94.5}              & \multicolumn{1}{c}{92.7}                         \\ 

SPS \cite{huang2021stochastic}                                          & ResNet50                         & \multicolumn{1}{c}{88.7}              & \multicolumn{1}{c}{94.9}              & \multicolumn{1}{c}{92.7}                         \\
CAL \cite{rao2021counterfactual}                                     & ResNet101                         & \multicolumn{1}{c}{90.6}              & \multicolumn{1}{c}{95.5}              & \multicolumn{1}{c}{94.2}     \\

\midrule
TransFG \cite{he2021transfg}                                      & ViT-Base                         & \multicolumn{1}{c}{91.7}              & \multicolumn{1}{c}{94.8}              & \multicolumn{1}{c}{-}       \\     
RAMS-Trans  \cite{hu2021rams}                                      & ViT-Base                         & \multicolumn{1}{c}{91.3}              & \multicolumn{1}{c}{-}              & \multicolumn{1}{c}{-}       \\     
FFVT \cite{wang2021feature}                                    & ViT-Base                         & \multicolumn{1}{c}{91.6}              & \multicolumn{1}{c}{-}              & \multicolumn{1}{c}{-} 
\\ 
\midrule
Baseline                                     & DeiT-Tiny                     & \multicolumn{1}{c}{82.1}              & \multicolumn{1}{c}{87.2}              & \multicolumn{1}{c}{84.7}                            \\
Baseline + DCAL            & DeiT-Tiny                     & \multicolumn{1}{c}{84.6}     & \multicolumn{1}{c}{89.4}     & \multicolumn{1}{c}{87.4}                         \\
Baseline                                     & DeiT-Small                     & \multicolumn{1}{c}{85.8}              & \multicolumn{1}{c}{90.7}              & \multicolumn{1}{c}{88.1}                              \\
Baseline + DCAL               & DeiT-Small                     & \multicolumn{1}{c}{87.6}     & \multicolumn{1}{c}{92.3}     & \multicolumn{1}{c}{90.0}                          \\
Baseline                                     & DeiT-Base                     & \multicolumn{1}{c}{88.0}              & \multicolumn{1}{c}{92.9}              & \multicolumn{1}{c}{90.3}                             \\
Baseline + DCAL               & DeiT-Base                     & \multicolumn{1}{c}{88.8}     & \multicolumn{1}{c}{93.8}     & \multicolumn{1}{c}{92.6}                           \\
Baseline                                     & ViT-Base                     & \multicolumn{1}{c}{90.8}              & \multicolumn{1}{c}{92.5}              & \multicolumn{1}{c}{90.0}                          \\
Baseline + DCAL              & ViT-Base                     & \multicolumn{1}{c}{91.4}     & \multicolumn{1}{c}{93.4}     & \multicolumn{1}{c}{91.5}                           \\
Baseline                                     & R50-ViT-Base                     & \multicolumn{1}{c}{91.3}              & \multicolumn{1}{c}{94.0}              & \multicolumn{1}{c}{92.4}                 \\
Baseline + DCAL              & R50-ViT-Base                     & \multicolumn{1}{c}{92.0}     & \multicolumn{1}{c}{95.3}     & \multicolumn{1}{c}{93.3}                 \\
\bottomrule
\end{tabular}
}
\end{center}
\vspace{-5mm}
\caption{Performance comparisons in terms of top-1 accuracy on three standard FGVC benchmarks: CUB-200-2011, Stanford Cars and  FGVC-Aircraft.}
\label{fine-grained sota compare}
\end{table}

\subsection{Results on Fine-Grained Visual Categorization}

We evaluate our method on three standard FGVC benchmarks and compare with the state-of-the-art approaches in Table \ref{fine-grained sota compare}. Our method achieves competitive performance compared to the prior CNN-based and Transformer-based methods. Particularly, with the R50-ViT-Base backbone, DCAL reaches 92.0\%, 95.3\% and 93.3\% top-1 accuracy on CUB-200-2011, Stanford Cars and FGVC-Aircraft benchmarks, respectively. Table \ref{fine-grained sota compare} also shows our method can consistently improve different vision Transformer baselines on all the three benchmarks, e.g., surpassing the pure Transformer (DeiT-Tiny) by 2.2\% and the hybrid structure of CNN and Transformer (R50-ViT-Base) by 1.3\% on Stanford Cars. The results validate the compatibility of our method to different Transformer architectures. 

\textbf{Comparisons to Transformer-based Methods.} 
Our method performs on par with the recent Transformer variants on FGVC: TransFG \cite{he2021transfg}, RAMS-Trans \cite{hu2021rams}, FFVT \cite{wang2021feature}. These existing methods also select tokens based on aggregated attention responses. Differently, they continue to model the selected tokens by self-attention while we perform cross-attention between local query and global key-value vectors. Compared to self-attention in selected tokens, we can relate the high-response regions with not only themselves but also with other context outside of them. Besides, TransFG \cite{he2021transfg} uses overlapping patches and will largely increase training time and computation overhead, while we adopt the standard non-overlapping patch split method.



\textbf{Comparisons to CNN-based Methods.} 
(1) Existing region-based methods can be divided to two categories. Explicit localization methods (e.g, RACNN \cite{fu2017look}, MA-CNN \cite{zheng2017learning}, NTS-Net \cite{yang2018learning}, MGE-CNN \cite{zhang2019learning}) utilize attention / localization sub-network with ranking losses to mine object regions. Implicit localization methods (e.g., S3N \cite{ding2019selective}, TASN \cite{zheng2019looking}) use class activation map and Gaussian sampling to amplify object regions in the original image. Our GLCA adopts a different scheme to incorporate the local information with higher performance, e.g., +3.5\% over MGE-CNN on CUB. (2) Pair-wise learning is also applied for FGVC by interacting features (CIN \cite{gao2020channel}, API-Net \cite{zhuang2020learning}) or introducing confusion (PC \cite{dubey2018pairwise}, SPS \cite{huang2021stochastic}) between image pairs during training. Our motivation of PWCA is similar to \cite{dubey2018pairwise,huang2021stochastic} but we implement a different regularization method to alleviate overfitting. Our method surpasses these related pair-wise learning methods, e.g., +3.9\% over CIN and +5.1\% over PC on CUB.

\begin{table*}[t!] 
\begin{center}
\scalebox{0.9}{
\begin{tabular}{l|cc|cc|cc|cc}
\toprule
 \multirow{2}{*}{Method}  & \multicolumn{2}{c|}{VeRi-776} & \multicolumn{2}{c|}{MSMT17} &\multicolumn{2}{c|}{Market1501} & \multicolumn{2}{c}{DukeMTMC} \\
   &   mAP (\%) & R1 (\%) & mAP (\%) & R1 (\%)  & mAP (\%) & R1 (\%)  & mAP (\%) & R1 (\%)   \\ \midrule
SPReID \cite{kalayeh2018spreid} & - & - & - & - & 83.4 & 93.7 & 73.3 & 86.0 \\
PCB \cite{sun2018pcb} & - & - & - & - & 81.6 & 93.8 & 69.2 & 83.3 \\
MGN \cite{wang2018learning}                     & -             & -    & 52.1  & 76.9& 86.9           & 95.7          & 78.4          & 88.7         \\
SAN \cite{jin2020semantics}                    & 72.5          & 93.3  &55.7&79.2      & 88.0           & 96.1          & 75.7          & 87.9         \\
 ABDNet \cite{chen2019abd}                     & -             & -    &60.8 &82.3       & 88.3           & 95.6          & 78.6          & 89.0         \\
 HOReID \cite{wang2020high}                  & -             & -    &- & -       & 84.9           & 94.2          & 75.6          & 86.9         \\
 ISP \cite{zhu2020identity}                & -             & -      &- & -      & 88.6           & 95.3          & 80.0          & 89.6         \\
STNReID \cite{luo2020stnreid} & - & - &- & -  & 84.9 & 93.8 & - & -             \\
CDNet \cite{li2021combined} & - & - & 54.7 & 78.9  & 86.0 & 95.1 & 76.8 & 88.6             \\
FIDI \cite{yan2021beyond} & 77.6 & 95.7 & - & -  & 86.8 & 94.5 & 77.5 & 88.1             \\
SPAN \cite{chen2020orientation}              & 68.9          & 94.0    &- & -      & -               & -                & -               & -              \\
PVEN \cite{meng2020parsing}                & 79.5          & 95.6     &- & -     & -               &  -             & -               &  -             \\
CAL (ResNet50) \cite{rao2021counterfactual}           & 74.3          & 95.4    &  56.2 & 79.5     &  87.0    &  94.5             &  76.4       & 87.2           \\   

\midrule
 DRL-Net \cite{jia2021drl} &-  & - & 55.3 & 78.4 & 86.9 & 94.7 & 76.6 & 88.1  \\ 
AAformer \cite{zhu2021aaformer} & - & - &63.2 & 83.6 &87.7 & 95.4 & 80.0 & 90.1 \\
 TransReID* (ViT-Base) \cite{he2021transreid} & 79.2 & 96.9 & 63.6 & 82.5 & - & - & - & - \\  \midrule
DeiT-Tiny              & 71.3 & 94.3 & 42.1 & 63.9 & 77.9 & 90.3 & 69.5 & 82.9             \\
DeiT-Tiny + DCAL (Ours)  & 74.1 & 94.7 & 44.9 & 68.2 & 79.8  & 91.8 & 71.7 & 84.9  \\ 
{DeiT-Small}   & 76.7  & 95.5 & 53.3 & 75.0 & 84.3 & 93.7  & 75.7 & 87.6             \\
DeiT-Small + DCAL (Ours)  & 78.1 & 95.9 & 55.1 & 77.3 & 85.3  & 94.0 & 77.4 & 87.9  \\ 
{DeiT-Base}          & 78.3 & 95.9   & 60.5 & 81.6 & 86.6  & 94.4  & 79.1 & 88.7             \\
DeiT-Base + DCAL (Ours)   & 80.0 & 96.5 & 62.3 & 83.1 & 87.2  & 94.5 & 80.2 & 89.6  \\ 
{ViT-Base}            & 78.1 & 96.0 & 61.6 & 81.4 & 87.1 & 94.3 & 78.9 & 89.4         \\
ViT-Base + DCAL (Ours)   & 80.2 & 96.9 & 64.0 & 83.1 & 87.5 & 94.7 & 80.1 & 89.0 \\ \bottomrule
\end{tabular}
}
\vspace{-1mm}
\caption{Performance comparisons on four Re-ID benchmarks: VeRi-776, MSMT17, Market1501, DukeMTMC. The input size is 256$\times$128 for pedestrian datasets and 256$\times$256 for vehicle datasets. * means results without side information for fair comparison.}
\label{reid sota compare}
\end{center}
\end{table*}

\subsection{Results on Object Re-ID}

We evaluate our method on four standard Re-ID benchmarks in Table \ref{reid sota compare} and achieve competitive performance compared to the state-of-the-art methods on both Person Re-ID and Vehicle Re-ID tasks. Particularly, with the ViT-Base backbone, DCAL reaches 80.2\%, 64.0\%, 87.5\%, 80.1\% mAP on VeRi-776, MSMT17, Market1501, DukeMTMC, respectively. Similar to FGVC, our method can consistently improve different vision Transformer baselines, e.g., surpassing the light-weight Transformer (DeiT-Tiny) by 2.8\% and the larger Transformer (ViT-Base) by 2.4\% on MSMT17.

\textbf{Comparisons to Transformer-based Methods.} 
Our method performs on par with the recent Transformer variants on Re-ID: DRL-Net \cite{jia2021drl}, AAformer \cite{zhu2021aaformer}, TransReID \cite{he2021transreid}. DRL-Net \cite{jia2021drl} imposes decorrelation constraints on Transformer decoder to disentangle ID relevant and irrelevant features, while we only employ Transformer encoder and extend self-attention to cross-attention. Both of existing methods (TransReID \cite{he2021transreid}, AAformer \cite{zhu2021aaformer}) and our methods incorporate local information for recognition but adopt different manners. TransReID \cite{he2021transreid} designs a jigsaw patch module to shuffle the patch embeddings for learning robust features. AAformer \cite{zhu2021aaformer} computes the attention between a part token and its associated subset of patch embeddings by online clustering. Differently, we proposes global-local cross-attention to enhance the interactions between global images and local regions. 

\textbf{Comparisons to CNN-based Methods.} (1) Many prior approaches have been presented to encode discriminative part-level features for recognition. Typical part-based ReID methods include SPReID \cite{kalayeh2018spreid} and PCB \cite{sun2018pcb}. SPReID \cite{kalayeh2018spreid} utilizes a parsing model to generate human part masks to compute reliable part representations, which consumes extra computation overhead in segmentation part. PCB \cite{sun2018pcb} utilizes a refined part pooling to retrieve the body part information. Our method does not aim to mine precise object parts but establish the interactions between global images and high-response local regions. (2) Image pairs or triplets are widely used in Re-ID for metric learning. Recent Re-ID methods also introduce pair-wise spatial transformer to match the holistic and partial image pairs \cite{luo2020stnreid} or design pair-wise loss to learn fine-grained features for recognition \cite{yan2021beyond}. Our pair-wise cross-attention is a new practice in Re-ID in contrast to previous work.


\begin{table*}[!t]
\begin{center}
\begin{tabular}{l|ccc|cccc|cccc}
\toprule
\multirow{2}{*}{Method} & \multicolumn{3}{c|}{CUB-200-2011} & \multicolumn{4}{c|}{VeRi-776}  & \multicolumn{4}{c}{MSMT17} \\
  & Params & FLOPs & Acc  & Params & FLOPs  & mAP & R1  & Params & FLOPs & mAP  & R1    \\ 
\midrule
Baseline    &5.5M &8.6G & 82.1 & 81.6M & 41.1G & 78.1  & 96.0 & 81.6M &20.5G  &61.6 & 81.4     \\
+ GLCA       &6.0M &8.8G & 83.1 & 88.4M& 42.4G & 79.5  & 96.5 & 88.4M &21.3G & 63.7 & 83.0       \\
+ PWCA       &5.5M &8.6G & 83.1 &81.6M & 41.1G & 79.2  & 96.5 &  81.6M &20.5G &62.8 &82.3       \\
Ours        &6.0M &8.8G & 84.6 & 88.4M& 42.4G & 80.2  & 96.9  & 88.4M &21.3G &64.0 &83.1       \\ 
\bottomrule
\end{tabular}
\caption{Effect of the proposed two types of cross-attention learning on CUB-200-2011, VeRi-776 and MSMT17. We use DeiT-Tiny for CUB, ViT-Base for VeRi-776 and MSMT17 as baselines in this ablation experiment.}
\label{table:glca_pwca_module}
\end{center}
\end{table*}

\subsection{Ablation Study}

\textbf{Contributions from Algorithmic Components.} We examine the contributions from the two types of cross-attention modules using different vision Transformer baselines in Table \ref{table:glca_pwca_module}. We use DeiT-Tiny for FGVC and ViT-Base for Re-ID. With either GLCA or PWCA alone, our method can obtain higher performance than the baselines. With both cross-attention modules, we can further improve the results. We note that PWCA will be removed for inference so that it does not introduce extra parameters or FLOPs. We uses one GLCA module in our method, which only requires a small increase of parameters or FLOPs compared to the baseline.

\textbf{Ablation Study on GLCA.} (1) Cross-ViT \cite{chen2021crossvit} is a most recent method based on cross-attention for general image classification. It constructs two Transformer branches to handle image tokens of different sizes and uses the class token from one branch to interact with patch tokens from another branch. We implement this idea using the same selected local queries and the same DeiT-Tiny backbone. The cross-token strategy obtains 82.1\% accuracy on CUB, which is worse than our GLCA by 1\%. (2) Another possible baseline to incorporate local information is computing the self-attention for the high-response local regions (i.e., local query, key and value vectors). This local self-attention baseline obtains 82.6\% accuracy on CUB using the DeiT-Tiny backbone, which is also worse than our GLCA (83.1\%). (3) We conduct more ablation experiments to examine the effect of GLCA. We obtain 82.6\% accuracy on CUB by selecting local query randomly and obtain 82.8\% by selecting local query based on the penultimate layer only. Our GLCA outperforms both baselines, validating that mining high-response local query with aggregated attention map is effective for our cross-attention learning.


\begin{table}[t!]
\begin{center}
\scalebox{0.95}{
\begin{tabular}{l|c|c}
\toprule
\multirow{2}{*}{Method}   & CUB & MSMT17 \\
 & Acc & mAP \\
\midrule
Baseline                    & 82.1 &  61.6  \\
+ PWCA                      & 83.1 & 62.8      \\
+ Adding noise in $I_1$     & 77.3 & 56.0    \\
+ Adding noise in label of $I_1$   & 81.6 & 60.8   \\
+ $I_2$ from noise          & 82.1 & 62.1     \\
+ $I_2$ from COCO           & 82.5 & 62.2    \\
+ $I_2$ from intra-class only & 81.7 & 62.2 \\
+ $I_2$ from inter-class only   &  83.0 &  62.7  \\
+ $I_2$ from intra- \& inter-class (1:1) &  83.0 & 62.5 \\
\bottomrule
\end{tabular}
}
\caption{Comparisons of different regularization methods. DeiT-Tiny is used for CUB and ViT-Base is used for MSMT17. }
\label{regularization methods.}
\end{center}
\end{table}

\textbf{Ablation Study on PWCA.} 
We compare PWCA with different regularization strategies in Table \ref{regularization methods.} by taking $I_1$ as the target image. The results show that adding image noise or label noise without cross-attention causes degraded performance compared to the self-attention learning baseline. As the extra image $I_2$ used in PWCA can be viewed as distractor, we also test replacing the key and value embeddings of $I_2$ with Gaussian noise. Such method performs better than adding image / label noise, but still worse than our method. Moreover, sampling $I_2$ from a different dataset (i.e., COCO), sampling intra-class / inter-class pair only, or sampling intra-class \& inter-class pairs with equal probability performs worse than PWCA. We assume that the randomly sampled image pairs from the same dataset (i.e., natural distribution of the dataset) can regularize our cross-attention learning well.

\begin{figure}[t!]
\begin{center}
\begin{tabular}{@{}cc@{}}
\includegraphics[width = 0.45\linewidth]{{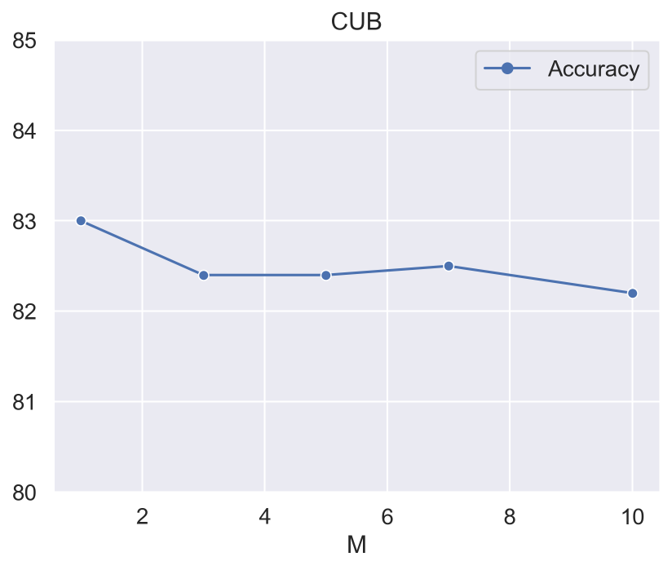}} & 
\includegraphics[width = 0.45\linewidth]{{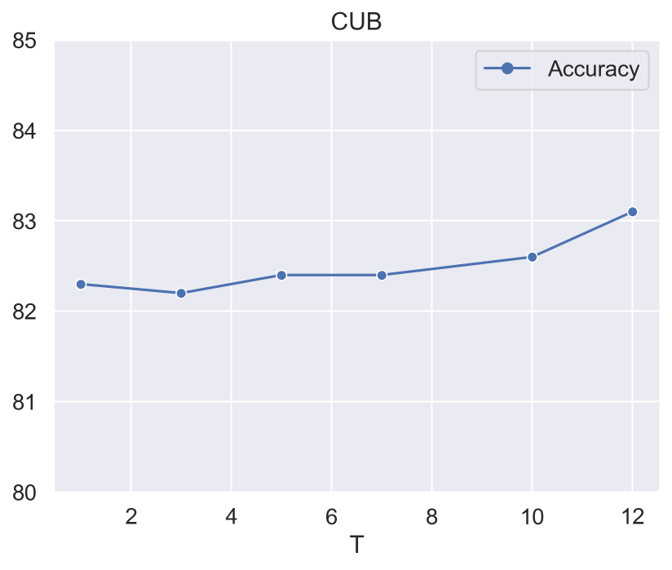}} \\
\end{tabular}
\begin{tabular}{@{}cc@{}}
\includegraphics[width = 0.45\linewidth]{{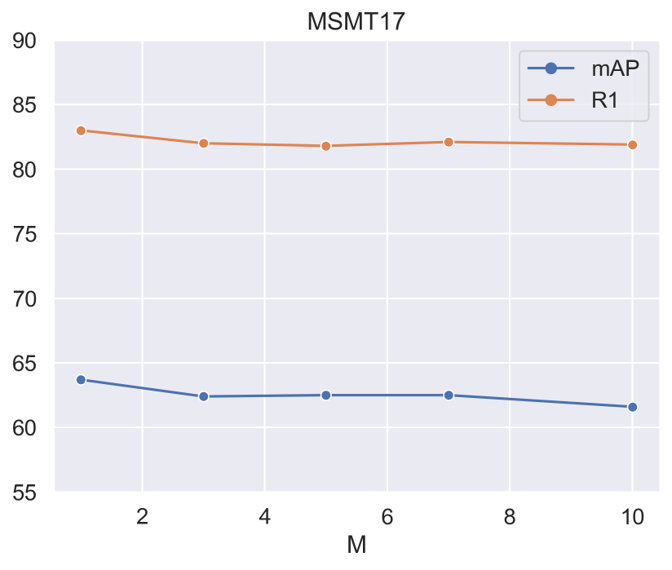}} & 
\includegraphics[width = 0.45\linewidth]{{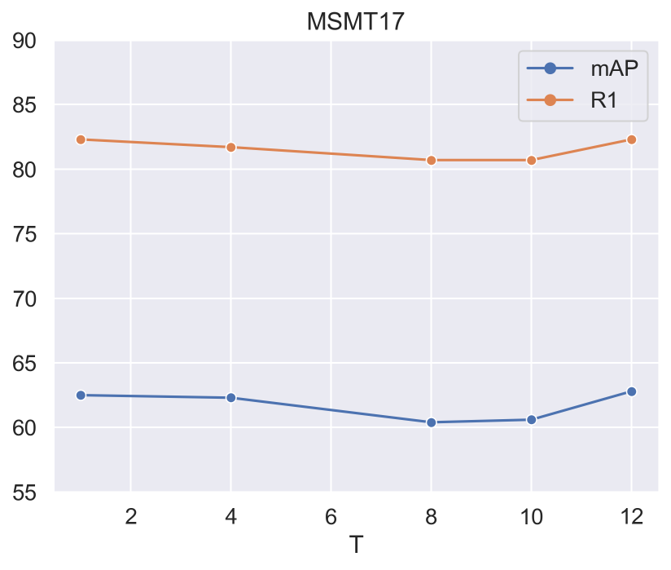}}\\
\end{tabular}
\end{center}
\vspace{-5mm}
\caption{Effect on the amount of cross-attention blocks. DeiT-Tiny is used for CUB and ViT-base ise used for MSMT17. For all the backbones and all the datasets, we build the same $M=1$ GLCA block and same $T=12$ PWCA blocks in our method.}
\label{figure:block}
\end{figure}

\textbf{Amount of Cross-Attention Blocks.} Figure \ref{figure:block} presents the ablation experiments on the amount of our cross-attention blocks using DeiT-Tiny for CUB and ViT-Base for MSMT17. For GLCA, the results show that $M=1$ performs best. We analyze that the deeper Transformer encoder can produce more accurate accumulated attention scores as the attention flow is propagated from the input layer to higher layer. Moreover, using one GLCA block only introduces small extra Parameters and FLOPs for inference. For PWCA, the results show that $T=12$ performs best. It implies that adding $I_2$ throughout all the encoders can sufficiently regularize the network as our self-attention baseline has $L=12$ blocks in total. Note that PWCA is only used for training and will be removed for inference without consuming extra computation cost.

\begin{figure}[t!]
\begin{center}
\begin{tabular}{@{}c@{}}
\includegraphics[width = 0.8\linewidth]{{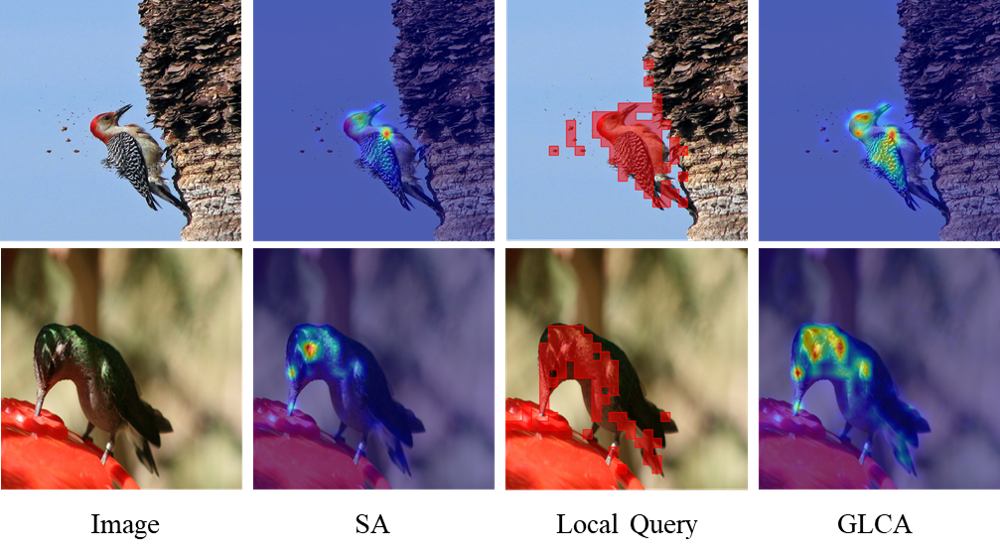}} \vspace{-2mm} \\
\footnotesize{(a) SA vs. GLCA} \\
\end{tabular}
\vspace{-2mm}
\begin{tabular}{@{}c@{}}
\includegraphics[width = 0.8\linewidth]{{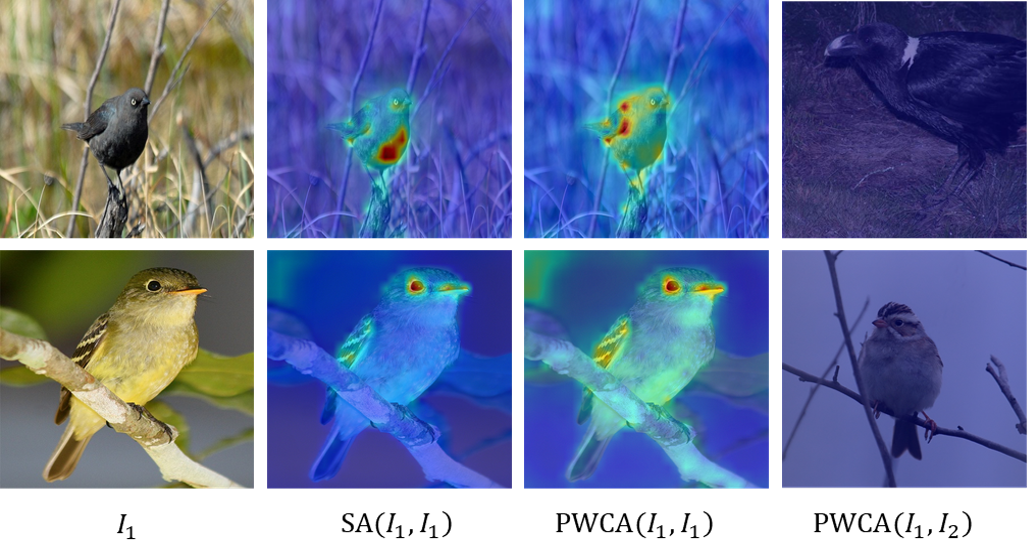}} \vspace{-2mm} \\
\footnotesize{(b) SA vs. PWCA} \\
\end{tabular}
\vspace{-3mm}
\end{center}
\caption{Visualization of the generated attention map for self-attention learning and our cross-attention learning on CUB.}
\label{figure: heatmap_cub}
\end{figure}

\begin{figure}[t!]
\begin{center}
\begin{tabular}{@{}cc@{}}
\includegraphics[width = 0.5\linewidth]{{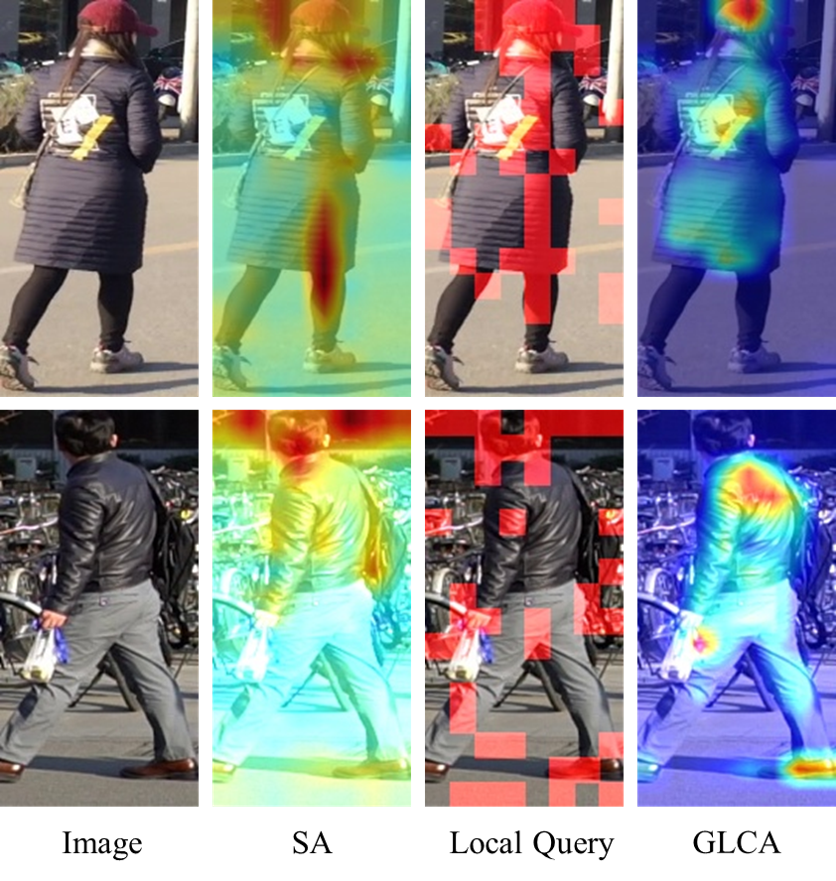}} & \hspace{-3mm} \includegraphics[width = 0.5\linewidth]{{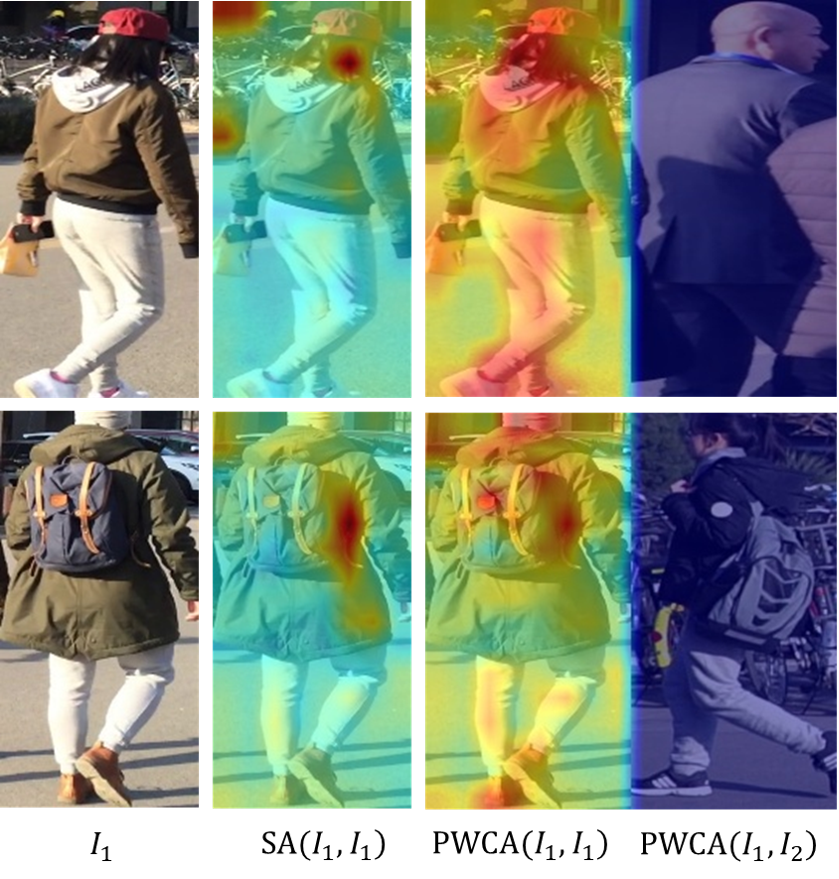}} \\
\footnotesize{(a) SA vs. GLCA} & \footnotesize{(b) SA vs. PWCA} \\
\end{tabular}
\end{center}
\vspace{-3mm}
\caption{Visualization of the generated attention map for self-attention learning and our cross-attention learning on MSMT17.}
\label{figure: heatmap_msmt}
\end{figure}

\subsection{Qualitative Analysis}
Figure \ref{figure: heatmap_cub} (a) and Figure \ref{figure: heatmap_msmt} (a) visualize the generated attention map using \cite{abnar2020quantifying} and the selected high-response patches. We observe that self-attention tend to highlight the most discriminative regions in the image. Thanks to GLCA, our method can reduce misleading attention and encourage the network to discover more discriminative clues for recognition.

Figure \ref{figure: heatmap_cub} (b) and Figure \ref{figure: heatmap_msmt} (b) visualize the generated attention map using \cite{abnar2020quantifying} for self-attention and PWCA. We observe that PWCA can diffuse the attention responses to explore more complementary parts of objects compared to self-attention. We also visualize the attention map on the distractor image and the blue gauze on it indicates that little attention is derived. It is accordance with our expectation that the attention weights will dominate on the target image as we compute the cross-attention between the query of target image and the combined key-value vectors (Eq. \ref{eq:pwca}).

\subsection{Limitations}
Compared to the self-attention learning baseline, our method may take longer time for network convergence as we perform joint training of self-attention and the proposed two types of cross-attention. For example, the self-attention baseline costs 2.1 hours while our method costs 3.8 hours for training on CUB with the same DeiT-backbone and same epochs of 100. However, it is noted that fine-grained recognition datasets are much smaller than the large-scale image classification benchmark and thereby our training time in practice is still acceptable.

Another limitation is that GLCA will increase small computation cost compared to the self-attention baseline. For example, Table \ref{table:glca_pwca_module} shows that GLCA increases 9\% Params and 2\% FLOPs for DeiT-Tiny on CUB and increases 8\% Params and 3\% FLOPs for ViT-Base on VeRi-776. We also test removing both GLCA and PWCA blocks for maintaining the same computation cost with the self-attention baseline, and the performance slightly drops, e.g, 84.3\% vs. 84.6\% (Ours) accuracy on CUB and 80.1\% vs. 80.2\% (Ours) mAP on VeRi-776.
\section{Conclusion}
In this work, we introduce two types of cross-attention mechanisms to better learn subtle feature embeddings for recognizing fine-grained objects. GLCA can help reinforce the spatial-wise discriminative clues by modeling the interactions between global images and local regions. PWCA can establish the interactions between image pairs and can be viewed as a regularization strategy to alleviate overfitting. Our cross-attention design is easy-to-implement and compatible to different vision Transformer baselines. Extensive experiments on seven benchmarks have demonstrated the effectiveness of our method on FGVC and Re-ID tasks. We expect that our method can inspire new insights for the self-attention learning regime in Transformer.

\clearpage
\appendix

\section{Overview}
In this supplementary material, we present more experimental results and analysis.
\begin{itemize}
    \item We test different inference architectures. 
    \item We provide additional ablation study on effect of ratio of local query selection.
    \item We show more visualization results of generated attention maps on different benchmarks. 
    \item We conduct experiments on more Transformer baselines.
\end{itemize}

\section{Different Inference Architectures}
Our default inference architecture is that all the PWCA modules are removed and only SA and GLCA modules are used. For FGVC, we add class probabilities output by classifiers of SA and GLCA for prediction. For Re-ID, we concat two final class tokens of SA and GLCA as the output feature for prediction. We also test two different inference architectures: (1) ``SA'': using the last SA module for inference. (2) ``GLCA'': using the GLCA module for inference. Table \ref{table:fine-grained} and \ref{table:reid} present the detailed performance with different baselines on all the FGVC and Re-ID benchmarks, respectively. The results show that only using the SA or GLCA module can obtain similar performance with our default setting. It is also noted that ``SA'' has the same inference architecture with the baseline by removing all the PWCA and GLCA modules for inference, which does not introduce extra computation cost.

\section{Ablation Study on Effect of $R$}
\begin{figure}[t!]
\begin{center}
\begin{tabular}{@{}cc@{}}
\includegraphics[width = 0.45\linewidth]{{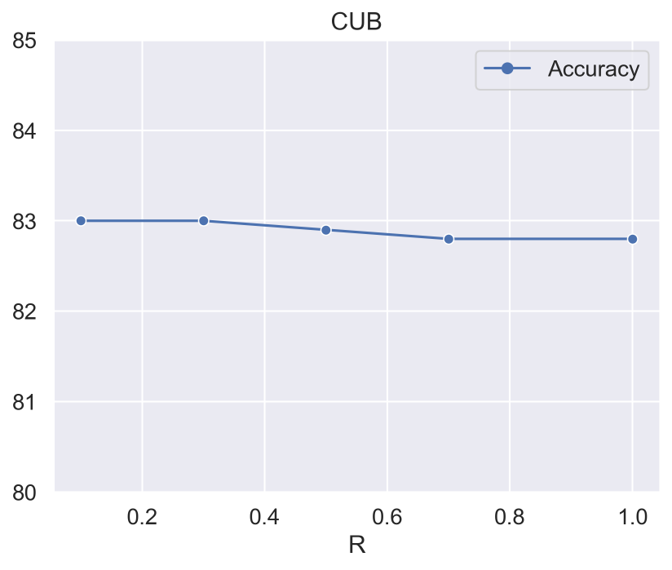}} & 
\includegraphics[width = 0.45\linewidth]{{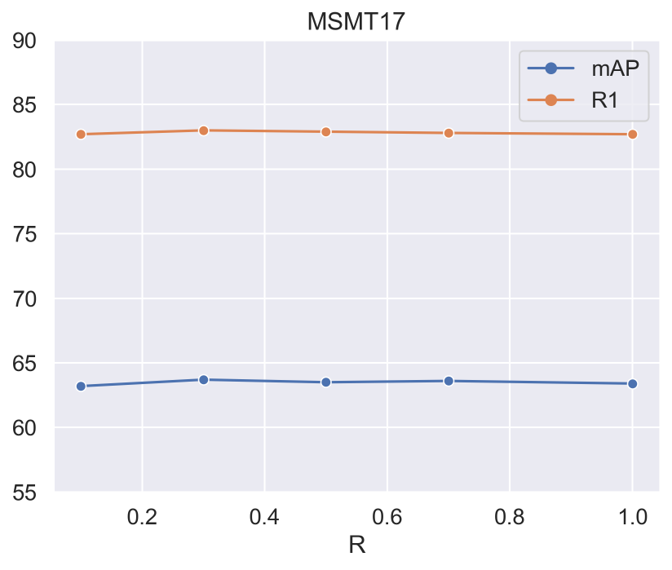}} \\
\end{tabular}
\end{center}
\vspace{-5mm}
\caption{Effect on the ratio of local query selection. DeiT-Tiny is used for CUB and ViT-base is used for MSMT17. We set $R=10\%$ for all the FGVC benchmarks and set $R=30\%$ for all the Re-ID benchmarks as default in our method.}
\label{figure:r}
\end{figure}

We test different choices of the ratios of selecting high-response regions as local query. Figure \ref{figure:r} shows that different choices of $R$ can obtain similar performance. We set $R=10\%$ for all the FGVC benchmarks and set $R=30\%$ for all the Re-ID benchmarks as default in our method. 

\section{More Visualization Results}
\begin{figure}[t!]
\begin{center}
\begin{tabular}{@{}c@{}}
\includegraphics[width = 0.8\linewidth]{{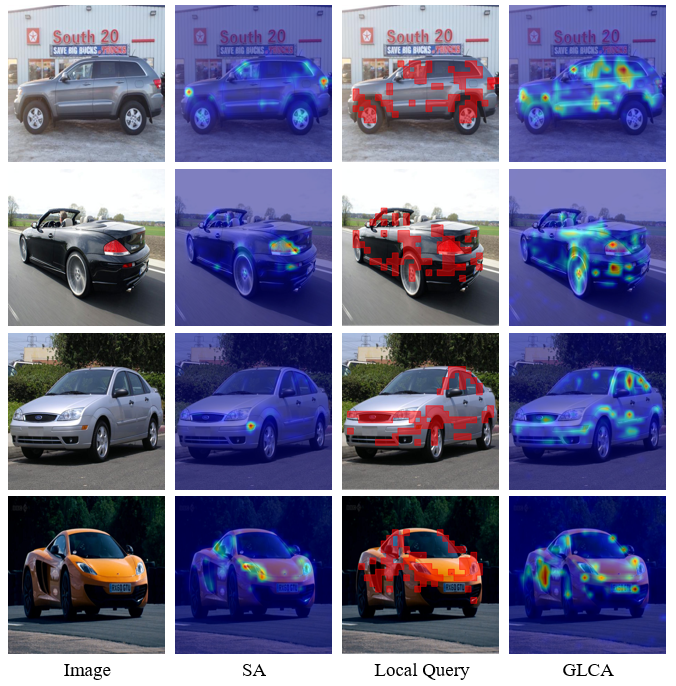}}  \\
\vspace{2mm}
\footnotesize{(a) SA vs. GLCA} \\
\end{tabular}
\vspace{-2mm}
\begin{tabular}{@{}c@{}}
\includegraphics[width = 0.8\linewidth]{{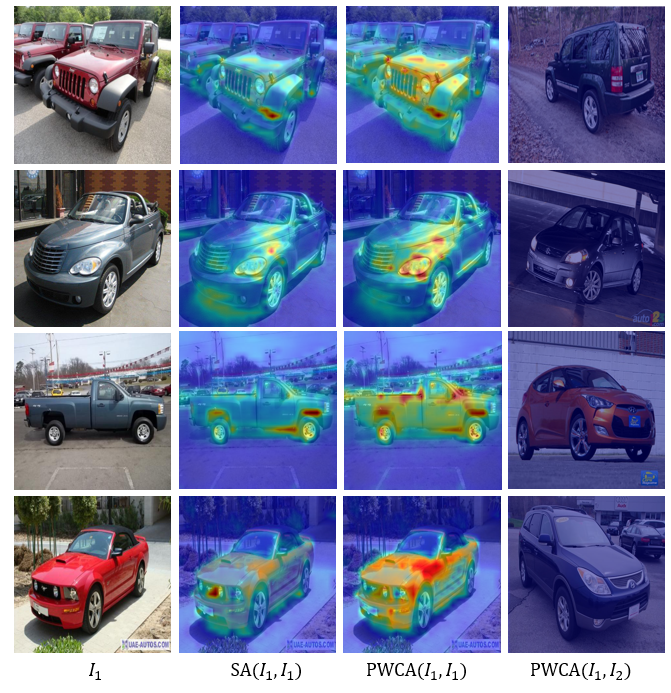}} \\
\footnotesize{(b) SA vs. PWCA} \\
\end{tabular}
\vspace{-2mm}
\end{center}
\caption{Visualization of the generated attention map for self-attention learning and our cross-attention learning on Stanford-Cars.}
\label{figure: heatmap_car}
\end{figure}

\begin{figure}[t!]
\begin{center}
\begin{tabular}{@{}c@{}}
\includegraphics[width = 0.8\linewidth]{{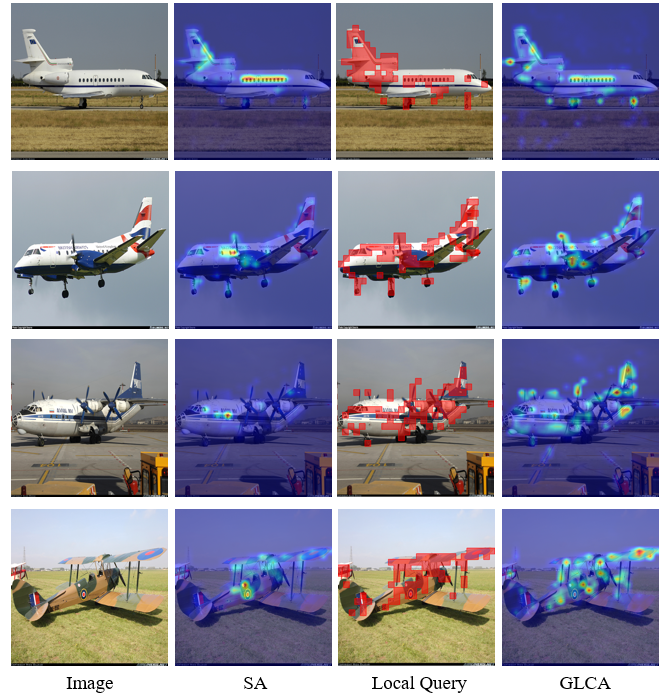}}  \\
\vspace{2mm}
\footnotesize{(a) SA vs. GLCA} \\
\end{tabular}
\vspace{-2mm}
\begin{tabular}{@{}c@{}}
\includegraphics[width = 0.8\linewidth]{{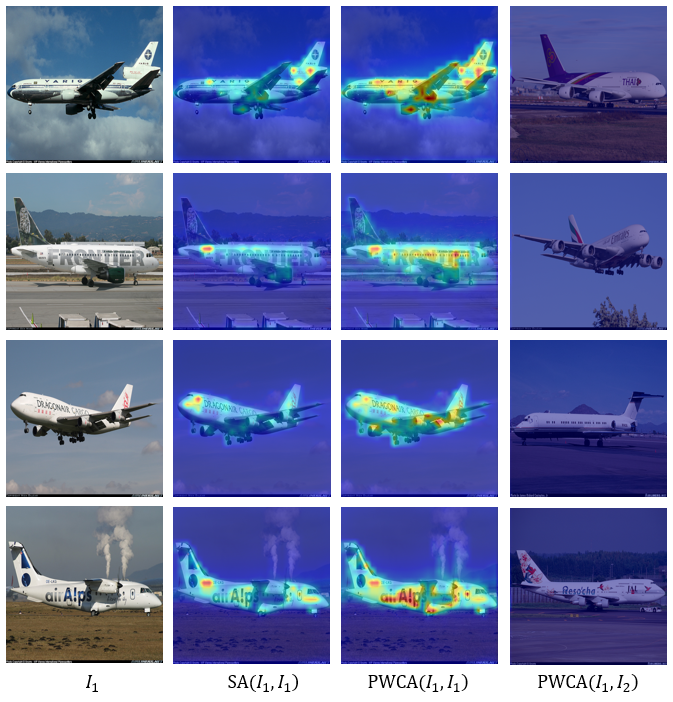}} \\
\footnotesize{(b) SA vs. PWCA} \\
\end{tabular}
\vspace{-2mm}
\end{center}
\caption{Visualization of the generated attention map for self-attention learning and our cross-attention learning on FGVC-Aircraft.}
\label{figure: heatmap_air}
\end{figure}

\begin{figure}[t!]
\begin{center}
\begin{tabular}{@{}c@{}}
\includegraphics[width = 0.8\linewidth]{{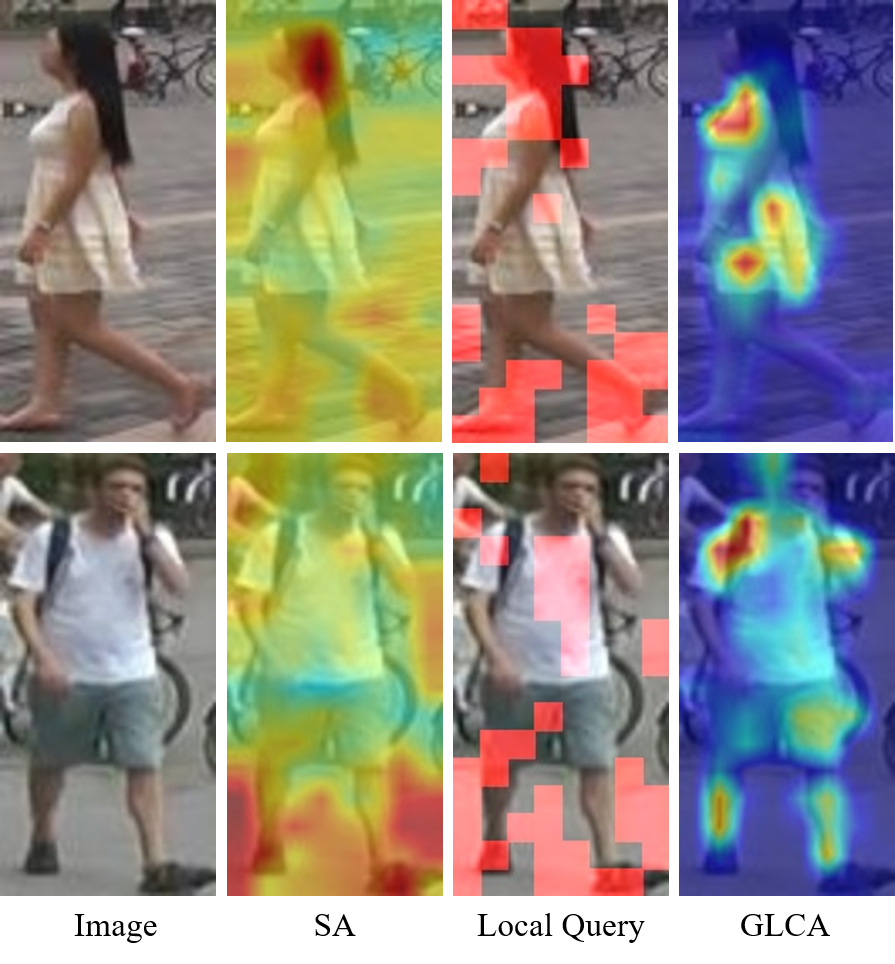}}  \\
\vspace{2mm}
\footnotesize{(a) SA vs. GLCA} \\
\end{tabular}
\vspace{-2mm}
\begin{tabular}{@{}c@{}}
\includegraphics[width = 0.8\linewidth]{{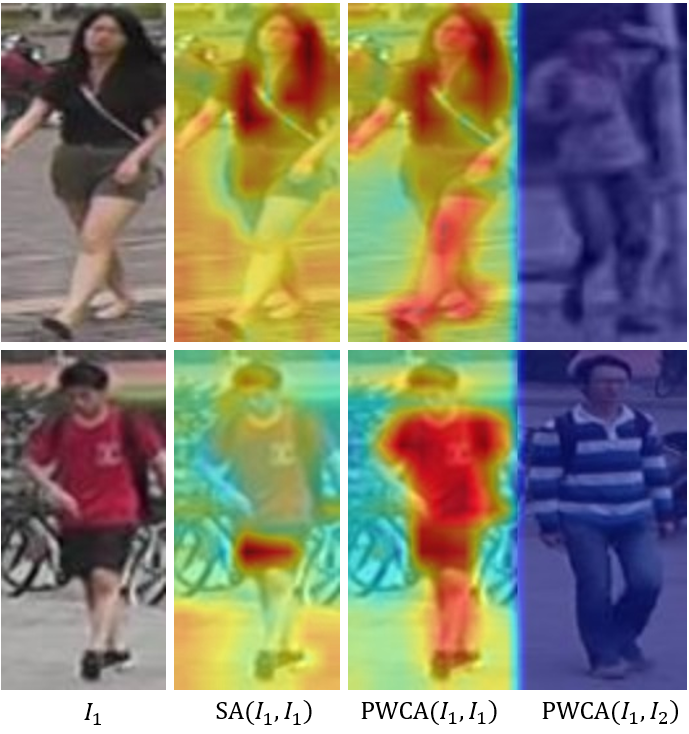}} \\
\footnotesize{(b) SA vs. PWCA} \\
\end{tabular}
\vspace{-2mm}
\end{center}
\caption{Visualization of the generated attention map for self-attention learning and our cross-attention learning on Market-1501.}
\label{figure: heatmap_market}
\end{figure}

\begin{figure}[t!]
\begin{center}
\begin{tabular}{@{}c@{}}
\includegraphics[width = 0.8\linewidth]{{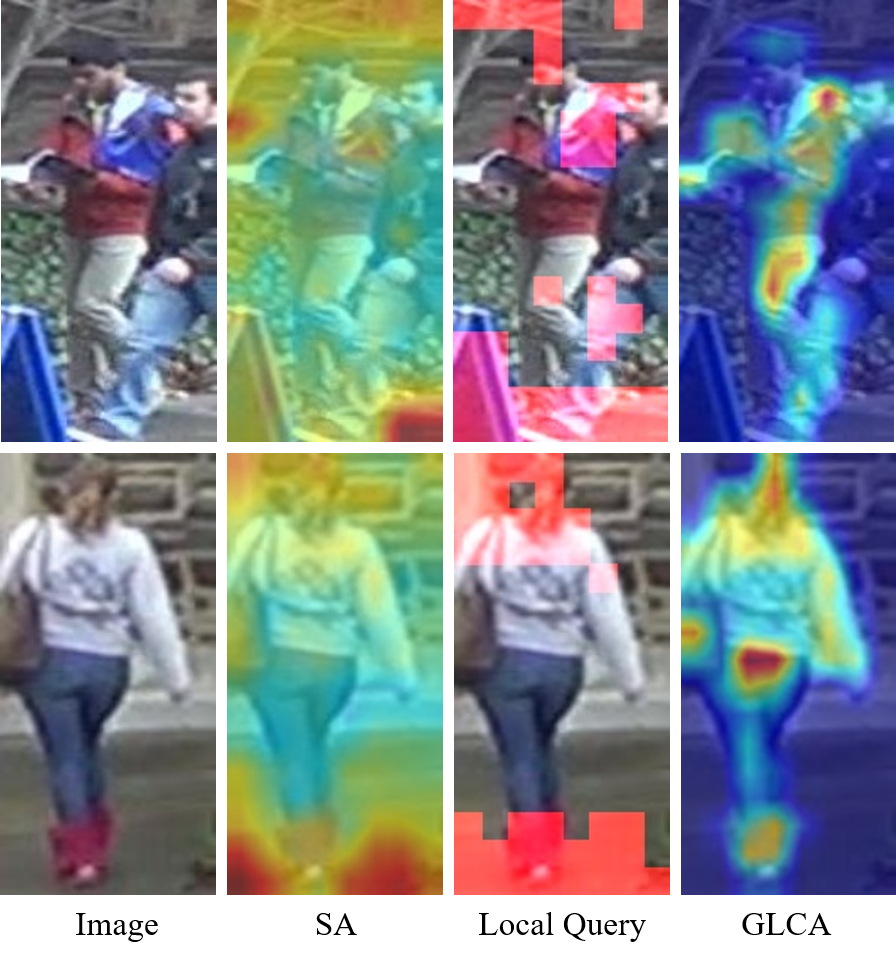}}  \\
\vspace{2mm}
\footnotesize{(a) SA vs. GLCA} \\
\end{tabular}
\vspace{-2mm}
\begin{tabular}{@{}c@{}}
\includegraphics[width = 0.8\linewidth]{{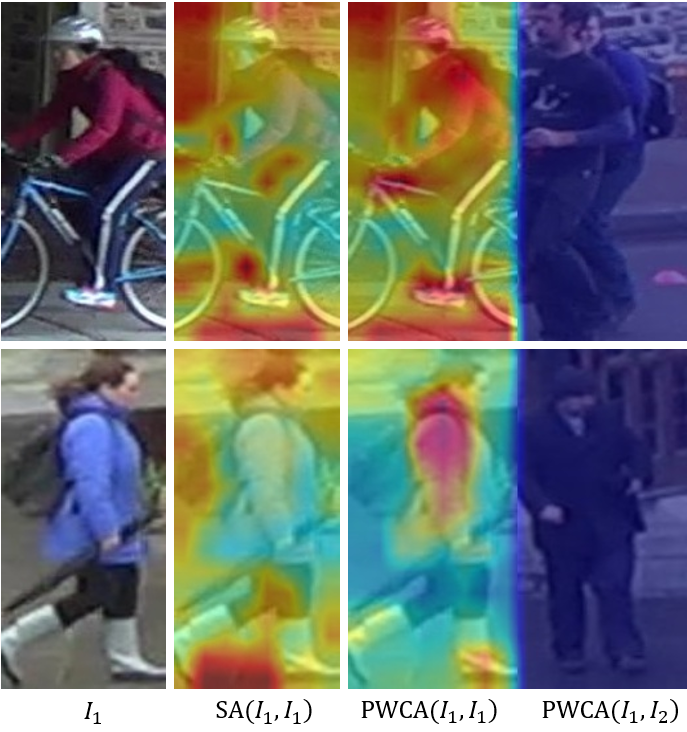}} \\
\footnotesize{(b) SA vs. PWCA} \\
\end{tabular}
\vspace{-2mm}
\end{center}
\caption{Visualization of the generated attention map for self-attention learning and our cross-attention learning on DukeMTMC-ReID.}
\label{figure: heatmap_duke}
\end{figure}

\begin{figure}[t!]
\begin{center}
\begin{tabular}{@{}c@{}}
\includegraphics[width = 0.8\linewidth]{{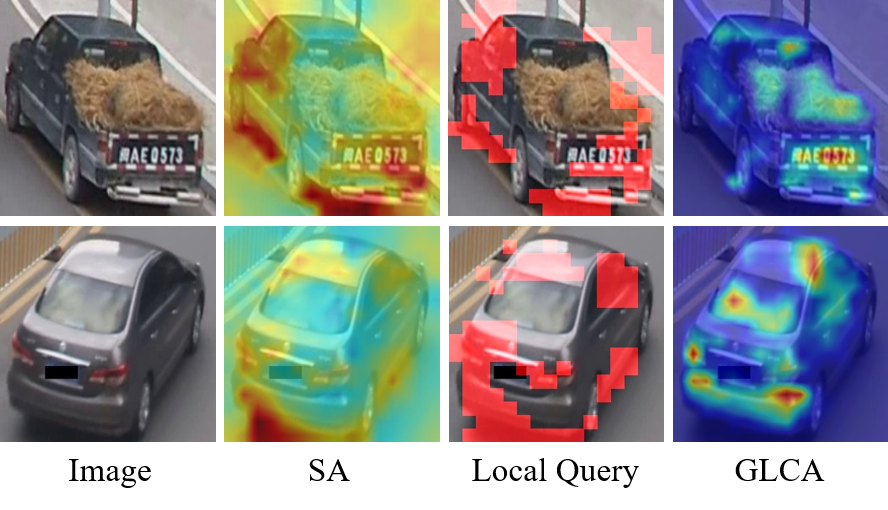}}  \\
\vspace{2mm}
\footnotesize{(a) SA vs. GLCA} \\
\end{tabular}
\vspace{-2mm}
\begin{tabular}{@{}c@{}}
\includegraphics[width = 0.8\linewidth]{{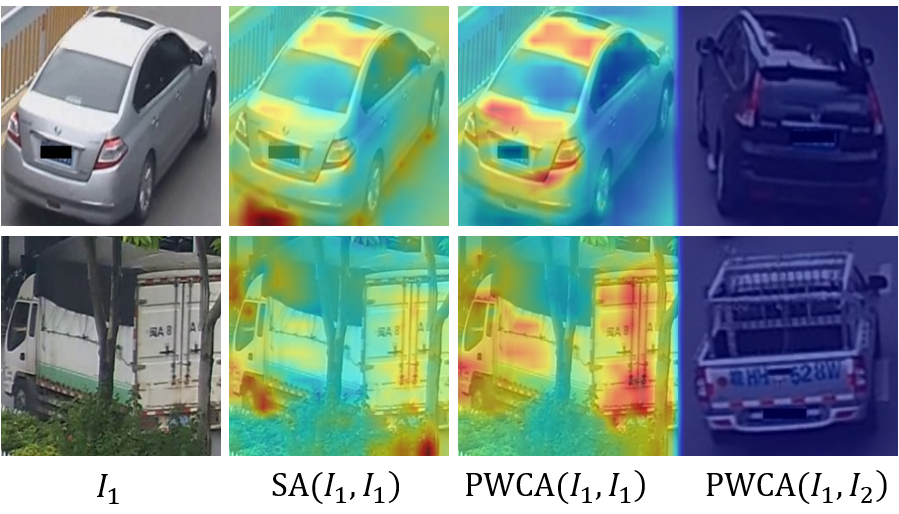}} \\
\footnotesize{(b) SA vs. PWCA} \\
\end{tabular}
\vspace{-2mm}
\end{center}
\caption{Visualization of the generated attention map for self-attention learning and our cross-attention learning on VeRi-776.}
\label{figure: heatmap_veri}
\end{figure}

We show more visualization results by comparing self-attention and our cross-attention method. Figure \ref{figure: heatmap_cub}, \ref{figure: heatmap_car}, \ref{figure: heatmap_air} present the generated attention maps on different FGVC benchmarks. Figure \ref{figure: heatmap_market}, \ref{figure: heatmap_duke}, \ref{figure: heatmap_veri} present the generated attention maps on different Re-ID benchmarks. The results show that our DCAL can reduce misleading attentions and diffuse the attention response to discover more complementary parts for recognition.

\section{More Transformer Baselines}
We conduct two more experiments on CaiT \cite{touvron2021going} and Swin Transformer \cite{liu2021swin}. CaiT-XS24 obtains 88.5\% while our method obtains 89.7\% top-1 accuracy on CUB. Swin-T obtains 84.9\% while our method obtains 85.8\% top-1 accuracy on CUB. For Re-ID on MSMT, Swin-T achieves 55.7\% while we achieve 56.7\% mAP. As locality has been incorporated by windows in Swin Transformer, we only apply PWCA into it. 

{\small
\bibliographystyle{ieee_fullname}
\bibliography{egbib}
}

\clearpage

\begin{table}[t!]
\scalebox{0.97}{
\begin{subtable}{0.5\textwidth} 
\begin{tabular}{c|ccc}
\toprule
Model       & SA (\%)   &GLCA (\%)  & SA+GLCA (\%)   \\ 
\midrule
DeiT-Tiny      &84.4      &83.6     & 84.6       \\
DeiT-Small    &87.6      & 87.4     & 87.6              \\
DeiT-Base    & 88.7      & 88.5     & 88.8       \\
ViT-Base   & 91.3      & 91.4     & 91.4             \\
R50-ViT-Base    & 91.5      & 91.9     & 92.0     \\
\bottomrule
\end{tabular}
\caption{CUB-200-2011}
\vspace{3mm}
\end{subtable}
}
\scalebox{0.97}{
\begin{subtable}{0.5\textwidth} 
\begin{tabular}{c|ccc}
\toprule
Model       & SA (\%)   &GLCA (\%)  & SA+GLCA (\%)   \\ 
\midrule
DeiT-Tiny      &89.2      &87.8     & 89.4       \\
DeiT-Small    & 92.4      & 91.8     & 92.3               \\
DeiT-Base    & 93.9      & 93.5     & 93.8       \\
ViT-Base   & 93.5      & 92.9     & 93.4             \\
R50-ViT-Base    & 95.3      & 94.8     & 95.3     \\
\bottomrule
\end{tabular}
\caption{Stanford-Cars}
\vspace{3mm}
\end{subtable}
}
\scalebox{0.97}{
\begin{subtable}{0.5\textwidth} 
\begin{tabular}{c|ccc}
\toprule
Model       & SA (\%)   &GLCA (\%)  & SA+GLCA (\%)   \\ 
\midrule
DeiT-Tiny      &86.9      &86.7     & 87.4        \\
DeiT-Small    &90.1      & 89.8     & 90.0              \\
DeiT-Base    & 92.5      & 92.3     & 92.6       \\
ViT-Base   & 91.4     & 91.1     & 91.5             \\
R50-ViT-Base    & 93.3      & 93.1     & 93.3      \\
\bottomrule
\end{tabular}
\caption{FGVC-Aircraft}
\end{subtable}
}
\caption{Ablation study on different inference architectures for FGVC in terms of accuracy. SA: using SA as the last layer to output class probabilities. GLCA: using GLCA as the last layer to output class probabilities. SA+GLCA: combine the output of SA and GLCA for inference.} 
\label{table:fine-grained}
\end{table}

\begin{table}[t!]
\scalebox{0.97}{
\begin{subtable}{0.45\textwidth} 
\begin{tabular}{c|ccc}
\toprule
Model & SA (\%)   &GLCA (\%)  & SA+GLCA (\%)  \\ 
\midrule
DeiT-Tiny      &44.8 / 68.1      &44.8 / 68.1    & 44.9 / 68.2       \\
DeiT-Small    & 54.9  / 77.4      & 55.1 / 77.2      & 55.1 / 77.3               \\
DeiT-Base    & 62.2 / 83.1      & 62.3 / 83.1    & 62.3 / 83.1       \\
ViT-Base   & 63.9 / 83.2     & 63.9 / 83.1    & 64.0 / 83.1             \\
\bottomrule
\end{tabular}
\caption{MSMT17}
\vspace{3mm}
\end{subtable}
}
\scalebox{0.97}{
\begin{subtable}{0.45\textwidth} 
\begin{tabular}{c|ccc}
\toprule
Model & SA (\%)   &GLCA (\%)  & SA+GLCA (\%)  \\ 
\midrule
DeiT-Tiny      &71.6 / 85.1      &71.7 / 84.9     & 71.7 / 84.9      \\
DeiT-Small    & 77.4 / 88.0      & 77.4 / 87.8     & 77.4 / 87.9              \\
DeiT-Base    & 80.2 / 89.9      & 80.2 / 89.6    & 80.2 / 89.6      \\
ViT-Base    &80.1 / 89.1      &80.1 / 89.0     & 80.1 / 89.0            \\
\bottomrule
\end{tabular}
\caption{DukeMTMC-ReID}
\vspace{3mm}
\end{subtable}
}
\scalebox{0.97}{
\begin{subtable}{0.45\textwidth} 
\begin{tabular}{c|ccc}
\toprule
Model  & SA (\%)   &GLCA (\%)  & SA+GLCA (\%) \\ 
\midrule
DeiT-Tiny      &79.7 / 91.8      &79.7 / 91.8     & 79.8 / 91.8      \\
DeiT-Small    & 85.2 / 94.1      &85.2 / 94.0     & 85.3 / 94.0              \\
DeiT-Base    & 87.2 / 94.5      & 87.2 / 94.4    & 87.2 / 94.5      \\
ViT-Base    &87.5 / 94.8      &87.5 / 94.7     & 87.5 / 94.7            \\
\bottomrule
\end{tabular}
\caption{Market1501}
\vspace{3mm}
\end{subtable}
}
\scalebox{0.97}{
\begin{subtable}{0.45\textwidth} 
\begin{tabular}{c|ccc}
\toprule
Model &SA (\%)   &GLCA (\%)  & SA+GLCA (\%) \\ 
\midrule
DeiT-Tiny      &74.1 / 94.6      &74.0 / 94.6     & 74.1 / 94.7      \\
DeiT-Small    & 78.0 / 95.9      &78.0 / 95.9     & 78.1 / 95.9          \\
DeiT-Base    & 79.9 / 96.6      & 80.0 / 96.6    & 80.0 / 96.5 \\
ViT-Base    &80.1 / 96.9     &80.2 / 96.9     & 80.2 / 96.9           \\
\bottomrule
\end{tabular}
\caption{VeRi-776}
\end{subtable}
}
\caption{Ablation study on different inference architectures for object Re-ID in terms of mAP and rank-1 accuracy. SA: using SA as the last layer to output final feature. GLCA: using GLCA as the last layer to output final feature. SA+GLCA: combine the output of SA and GLCA for inference.} 
\label{table:reid}
\end{table}

\begin{figure}[t!]
\begin{center}
\begin{tabular}{@{}c@{}}
\includegraphics[width = 0.8\linewidth]{{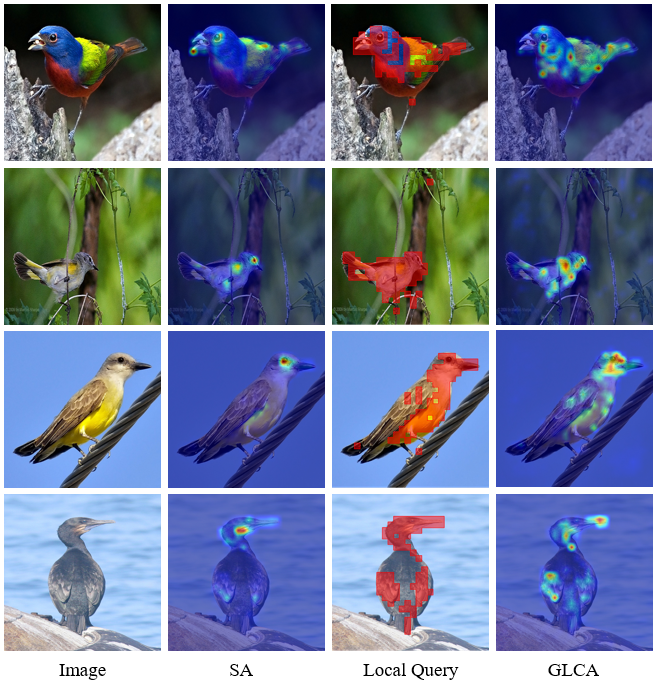}}  \\
\vspace{2mm}
\footnotesize{(a) SA vs. GLCA} \\
\end{tabular}
\vspace{-2mm}
\begin{tabular}{@{}c@{}}
\includegraphics[width = 0.8\linewidth]{{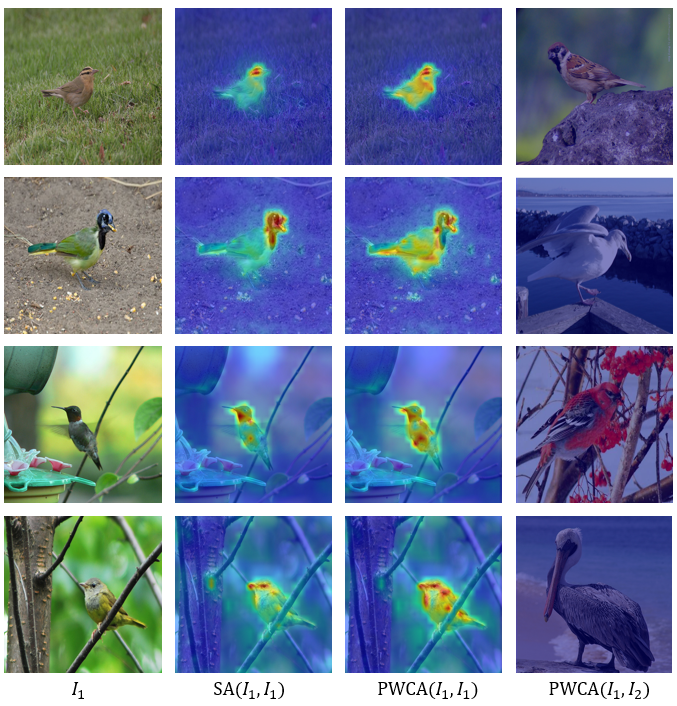}} \\
\footnotesize{(b) SA vs. PWCA} \\
\end{tabular}
\vspace{-2mm}
\end{center}
\caption{Visualization of the generated attention map for self-attention learning and our cross-attention learning on CUB-200-2011.}
\label{figure: heatmap_cub}
\end{figure}

\end{document}